\documentclass[11pt,a4paper]{article}
\usepackage[hyperref]{ranlp2021}
\usepackage{times}
\usepackage{latexsym}
\usepackage{graphicx}
\usepackage{scalefnt}
\usepackage{multirow}
\usepackage{booktabs}
\usepackage{amsmath}
\usepackage{lipsum}
\usepackage{tabularx}
\usepackage{comment}
\aclfinalcopy %remove number of lines in ACL template

\title{Contextual-Lexicon Approach for Abusive Language Detection}

\author{Francielle Vargas*†, Fabiana Góes*, Isabelle Carvalho‡ \\ \textbf{Fabrício Benevenuto†, Thiago A. S. Pardo*} \\
 \text{*}Institute of Mathematical and Computer Sciences, University of São Paulo, Brazil \\
 ‡Ribeirão Preto Medical School, University of São Paulo, Brazil \\
 †Computer Science Department, Federal University of Minas Gerais, Brazil\\
 
 \texttt{\{francielleavargas,fabianagoes,isabelle.carvalho,\}@usp.br}\\
\texttt{fabricio@dcc.ufmg.br, taspardo@icmc.usp.br}\\}

\begin{document}
\maketitle
\begin{abstract}
Since a lexicon-based approach is more elegant scientifically, explaining the solution components and being easier to generalize to other applications, this paper provides a new approach for offensive language and hate speech detection on social media. Our approach embodies a lexicon of implicit and explicit offensive and swearing expressions annotated with contextual information. Due to the severity of the social media abusive comments in Brazil, and the lack of research in Portuguese, Brazilian Portuguese is the language used to validate the models. Nevertheless, our method may be applied to any other language. The conducted experiments show the effectiveness of the proposed approach, outperforming the current baseline methods for the Portuguese language.\footnote{Proceedings of the 13th Recent Advances in Natural Language Processing, pages 1438–1447, 2021. \url{https://aclanthology.org/2021.ranlp-1.161/}}. %Moreover, the results suggests that the abusive vocabulary is ambiguous and highly context-dependent. 

\end{abstract}

%---------------------------------------------
\section{Introduction}
%---------------------------------------------
In Brazil, hate speech is prohibited. Nevertheless, in government and civil society, the regulation of hate speech is not effective due to the difficulty to identify, quantify and classify abusive comments. Indeed, this is rather a difficult requirement to satisfy.  According to \citeauthor{mesquita2018} (\citeyear{mesquita2018}), the \textit{Safernet} non-governmental organization, which operates in cooperation with public organizations in Brazil, as well as companies, such as Google, Facebook, and Twitter, proposed a collection of data on actions that violate human rights. The data is very worrisome: during the 2018 year's election period, denunciations with xenophobia content had an increase of 2,369.5\%; apology and public incitement to violence and crimes against life, 630.52\%; neo-nazism, 548.4\%; homophobia, 350.2\%; racism, 218.2\%; and religious intolerance, 145.13\% \footnote{\url{https://www.bbc.com/portuguese/brasil-46146756}}. Figure 1 shows the hate crimes evolution that occurred in the most populous Brazilian state \footnote{\url{https://www.ssp.sp.gov.br/}}. The data was collected from São Paulo public security government. The pink line provides data on religious intolerance crimes, red on homophobia/transphobia, blue on race/ethnicity/color, green on region/origin, yellow on political intolerance, and light green on other crimes. 

\begin{figure}[!ht]
\centering  
\includegraphics[width=0.49\textwidth]{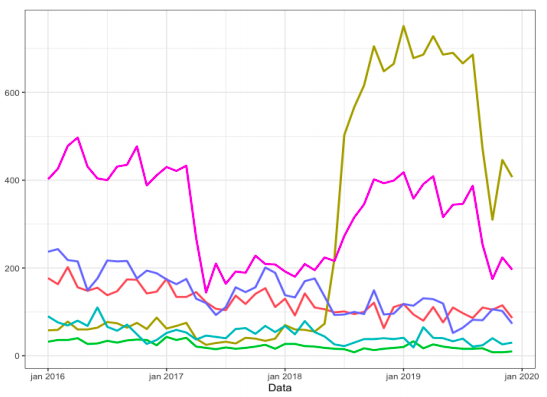}
\caption{Hate crimes occurrence in São Paulo from 2016 to beginning of 2020.}
\label{fig:hatecrimes}
\end{figure}

Indeed, it is generally agreed that the high incidence of hate crimes is boosted by the popularization of online social networks. In social media, people and organizations may use the language to defamation, oppression, and terrorism. The language used intentionally in order to disrespect, insult or attack the reader is denominated in literature by abusive language, unless otherwise stated such as offensive language \cite{coltekin2020corpus, pitenisetal2020offensive, Razavietall2010}, hate speech \cite{schmidtwiegand2017survey, waseemhovy2016hateful} and cyberbullying \cite{rosa2019}. 

According to \citeauthor{warnerhirschberg2012detecting} (\citeyear{warnerhirschberg2012detecting}), hate speech is a particular form of abusive language considering stereotypes to express an ideology of hate. In the same settings, \citeauthor{Nockleby2000} (\citeyear{Nockleby2000}) defines hate speech as ``any communication that disparages a person or a group based on some characteristic such as race, color, ethnicity, gender, sexual orientation, nationality, religion, or other characteristic''. %In \citeauthor{FortunaAndNunes2018} (\citeyear{FortunaAndNunes2018}), the authors claim that abusive language is an ``language that attacks or diminishes, that incites violence or hate against groups, based on specific characteristics such as physical appearance, religion, descent, national or ethnic origin, sexual orientation, gender identity or other, and it can occur with different linguistic styles, even in subtle forms or when humor is used''. 

To the best of our knowledge, no previous methods exist in order to embody an offensive lexicon annotated with contextual information to automatically classify abusive language on social media. Therefore, the main contribution of this paper is providing a new method for abusive comment detection on social media. Moreover, as already mentioned, due to the severity of the hate crimes in Brazil, and the lack of research in this language, Brazilian Portuguese is the language used to evaluate the proposed method, which showed high performance, outperforming the current baseline methods for Portuguese. Despite the proposed approach has been orchestrated over Brazilian Portuguese comments, the method in this paper may be applied to any other language. Finally, this paper also presents the evaluation of algorithms used for feature selection.

The remainder of the paper is structured as follows. In Section \ref{sec:related}, we briefly introduce the most relevant related work. Section \ref{sec:hatebr} presents a overview of the data. Sections \ref{sec:approach} and \ref{sec:experiments} describe the proposed method and the performed experiments. In Section \ref{subsec:results}, we report the evaluation results. In Section \ref{sec:conclusions}, we make some final remarks.

%---------------------------------------------
\section{Related Work}
\label{sec:related}
%---------------------------------------------

Several efforts have been made to provide automated detection approaches for hate speech and offensive languages on social media \cite{gaohuang2017detecting, davidsonetall2017, warnerhirschberg2012detecting}. The basic state of the art framework consists of creating lists of words that contain sets of known hate keywords. Furthermore, corpora are manually annotated in order to construct training datasets labeled with hate speech and non-hate speech. At last, automated methods of learning, such as traditional machine learning or neural-based machine learning, are used to automatically detect hate speech in social media texts. However, most hate speech resources and models are proposed for English \cite{zampierietal2019, basileetal2019semeval, davidsonetall2017, DennisEtAll2015, ting2013approach}. 

For Portuguese, \citeauthor{fortuna2019hierarchically} (\citeyear{fortuna2019hierarchically}) adopted the definition of hate speech proposed by \citeauthor{FortunaAndNunes2018} (\citeyear{FortunaAndNunes2018}), and proposed a new dataset composed of 5,668 tweets, as well as automated methods using a hierarchy of hate to identify social groups of discrimination. The authors have obtained 78\% f1-score using a neural network (LSTM). Additionally, \citeauthor{MoreiraAndPelle2017} (\citeyear{MoreiraAndPelle2017}) provide a new dataset composed of 1,250 comments collected from G1 Brazilian online newspaper \footnote{\url{https://g1.globo.com/}} and annotated with offensive and non-offensive tags. In addition, the authors present classification results achieved by classical machine learning algorithms (SMV and NB), reporting results over 81\% f1-score. 

%--------------------------------------------------
\section{Data Overview} 
\label{sec:hatebr}
%-------------------------------------------------
\subsection{HateBR Corpus}
%--------------------------------------------------
HateBR was proposed by \citeauthor{vargas2021annotating} (\citeyear{vargas2021annotating}), and consists of the first large-scale dataset for hate speech and offensive language detection for the Portuguese language. HateBR corpus annotation presents 75\% of human inter-annotator agreement. The corpus is composed of 7,000 Instagram comments annotated with three different layers: (i) binary classes (offensive and non-offensive); (ii) offense-levels (highly, moderately, and slightly offensive); and (iii) nine hate group targets (xenophobia, racism, homophobia, sexism, religious intolerance, partyism \footnote{According to the professor at Harvard University, ``partyism'' is a form of hostility and prejudice that operates across political lines \cite{Sunstein2016}. Moreover, \citeauthor{WestwoodEtAll2018} (\citeyear{WestwoodEtAll2018}) demonstrated that partyism influences behaviors and non-political judgments.}, apology to dictatorship, antisemitism, and fatphobia). 

The authors report that the comments were collected from six public Instagram accounts of the Brazilian political domain. Moreover, they selected three liberal-party accounts followed by three conservative-party accounts, being four women and two men. Due to the degree of complexity of the offensive language and hate speech detection tasks, mainly because it involves a highly politicized domain, the authors decided to enroll annotators at higher levels of education (Ph.D.), which are from different political orientations and colors in order to minimize bias. 

Tables \ref{tab:binary}, \ref{tab:granlevels}, \ref{tab:groupshate} show the HateBR dataset statistics. 

\begin{table}[!htb]
\caption{Binary classes: offensive x non-offensive.} 
\label{tab:binary}
\centering
\scalefont{0.75} %Scale o4f table font
\begin{tabular}{l|l}
\hline
\textbf{Binary Class} & \textbf{\begin{tabular}[c]{@{}l@{}}Total \\\end{tabular}} \\
\hline
Non-Offensive & 3,500 \\
Offensive & 3,500 \\
\hline
Total & 7,000 \\
\hline
\end{tabular}
\end{table}

\begin{table}[!htb]
\caption{Offense-level classes.} 
\label{tab:granlevels}
\centering
\scalefont{0.75} %Scale o4f table font
\begin{tabular}{l|l}
\hline
\textbf{Offense-levels Classes} & \textbf{\begin{tabular}[c]{@{}l@{}}Total \\ \end{tabular}}\\
\hline
Slightly Offensive & 1,281\\
Moderately Offensive & 1,440\\
Highly Offensive & 779\\
\hline
Total & 3,500\\
\hline
\end{tabular}
\end{table}
 
\begin{table}[!htb]
\caption{Hate group targets.}
\label{tab:groupshate}
\centering
\scalefont{0.75} %Scale o4f table font
\begin{tabular}{p{28mm}|p{7mm}}
\hline
\textbf{Hate Groups} &  \textbf{Total} \\
\hline
Partyism & 496\\
%\hline
Sexism & 97\\
%\hline
Religious Intolerance & 47\\
%\hline
Apology to Dictatorship & 32\\
%\hline
Fat Phobia & 27\\
%\hline
Homophobia & 17\\
%\hline
Racism & 8\\
%\hline
Antisemitism & 2\\
%\hline
Xenophobia & 1\\
\hline
\multicolumn{1}{l|}{Total} & 727\\
\hline
\end{tabular}
\end{table}

%--------------------------------------------------------
\subsection{MOL - Multilingual Offensive Lexicon}
\label{subsec:mol}
%--------------------------------------------------------
MOL (Multilingual Offensive Lexicon)\footnote{\url{https://github.com/francielleavargas/MOL}} consists of a multilingual offensive lexicon, composed of 1,000 explicit and implicit offensive and swearing expressions, which were annotated with a binary class: context-dependent and context-independent offensive. For example, the term \textit{`'vadia''} (``slut'') consists of a context-independent offensive term. On other hand,  the term \textit{inútil} (``useless'') is a context-dependent offensive term. Note that this last term is classified as context-dependent offensive because it also may be employed in a non-offensive context, such as ``this smartphone is useless'' or ``the process is useless for this task''. 

The MOL was extracted from HateBR corpus \cite{vargas2021annotating}, and each term or expression was annotated by three different annotators obtaining a high human agreement score (73\% Kappa). Furthermore, as already mentioned, implicit content also was extracted using ``clue terms or expressions''. For example, the expression \textit{voltar para a jaula} (``go back to the cage'') consists of a ``clue expression'' to identify the implicit offensive term \textit{ladrão} (``thief''). Finally, terms that showed explicit potential to indicate hate speech targets were also annotated, for instance, \textit{vadia} (``slut'') and \textit{judeus dos infernos} (``jews from hell''). Note that the occurrence of these cases may indicate sexist and antisemitism comments.

%--------------------------------------------
\section{The Proposed Approach}
\label{sec:approach}
%--------------------------------------------
%According to \citeauthor{kwoketaa2013} (\citeyear{kwoketaa2013}), bag-of-words approaches have provided high recall performance, however, it leads to high rates of false positives since the presence of offensive words can lead to the misclassification of tweets as hate speech. According to authors, in tweets that contain anti-black racism, eighty-six percent of the time was categorized as racist because it contained offensive words. Additionally, \citeauthor{wangetall2014} (\citeyear{wangetall2014}) argues offensive language on social media presents a relatively high prevalence of ``curse words'' becoming hate speech and offensive language detection particularly challenging. 

We present a new approach to detect abusive comments on the web and social media. Our method embodies an offensive lexicon, which provides contextual information on hate speech and offenses. We show in detail our approach in Sections 4.1, 4.2, and 4.3.

%----------------------------------------------------
\subsection{Tasks} 
\begin{figure}[!htb]
\centering  
\includegraphics[width=0.50\textwidth]{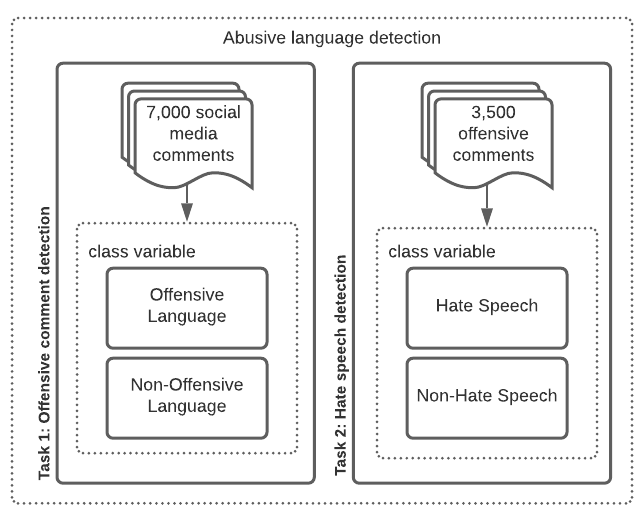}
\caption{Our approach to use the HateBR dataset and automatically detect offensive comments, as well as offensive comments with hate speech content.}
\label{fig:trainning}
\end{figure}

%---------------------------------------------------
In this paper, we assume that abusive language detection may be divided into two main tasks: (i) offensive language detection, (ii) hate speech detection. Considering this premise, we train two different classifiers. The first classifier automatically identifies offensive comments. On the other hand, the second classifier automatically identifies comments that present hate speech content. Note that a hate speech comment is always an offensive comment, however, an offensive comment may present or not hate speech content. Figure \ref{fig:trainning} shows each of these different tasks in detail.

%\begin{enumerate}
%  \item\textit{Offensive comments detection}: it consists of the binary classification in offensive class and non-offensive class;
%  \item \textit{Hate speech comments detection}: it consists of the classification of offensive class in hate speech and non-hate speech. 
%\end{enumerate} 

%--------------------------------------------------
\subsection{The Feature Set}
%--------------------------------------------------
Defining the most appropriate textual representation is a crucial task that directly influences the performance of the predictive model built by the classification algorithms. In this paper, we modeled hate speech and offensive language through different representation paradigms and features. We describe each one in what follows.

%--------------------------------------------------------
\subsubsection{Lexical and Morphosyntactic Features}
%--------------------------------------------------------
\label{subsubsec:lexical}
We selected lexical elements (each word into the document without stopwords), as well as part-of-speech-based features, using the Stanford Stanza POS tagger\footnote{\url{https://stanfordnlp.github.io/stanza/pos.html}} for Portuguese.

%--------------------------------------------------------
\subsubsection{Lexicon-Based Features}
\label{subsubsec:lexicon}
%-------------------------------------------------------
We included features from three different lexicons: one sentiment lexicon (Sentilex-PT \cite{sentilexpt2012}), one emotion lexicon (WordNetAffect.BR \cite{WordNetAffectBR2008}), and finally, one offensive contextual lexicon (MOL).

\begin{enumerate}
    \item \textbf{Sentiment Lexicon}: we evaluated features based on sentiment \cite{sentilexpt2012} and emotion \cite{WordNetAffectBR2008} lexicons, which present semantic polarity (e.g., positive, negative and neutral) and emotion types (e.g., anger, love, hate, disgust, suspicious and fear).
    
    \item \textbf{Contextual Lexicon}: we evaluated features based on an offensive lexicon (MOL) annotated with contextual (context-dependent and context-independent) labels. 
    
\end{enumerate}

%--------------------------------------------------
\subsubsection{Word Embedding Features}
\label{subsubsec:embedding}
%--------------------------------------------------
We also evaluated word embedding-based features. Different from other language models, BERT (Bidirectional Encoder Representations from Transformers) is usually used to pre-train deep bidirectional representations from unlabeled texts by jointly conditioning on both left and right contexts in all internal network layers \cite{bertdevlinetal2019}. In a similar setting, we also used fastText, the Facebook pre-trained models \cite{fasttex2016}. 

%--------------------------------------------------
\subsubsection{Feature Set Overview}
%--------------------------------------------------
We summarize in Table \ref{tab:features-models} the five feature representations used in this paper. 

\begin{table}[!htb]
\caption{Feature set representations.}
\label{tab:features-models}
\centering
\scalefont{0.75} %Scale o4f table font
\begin{tabular}{p{2mm}|p{20mm}|p{42mm}}
\hline
\textbf{N.} & \textbf{Features} &  \textbf{Description}\\
\hline
1& POS+S    & Bag-of-POS+Sentiment\\
\hline
2& BOW     & Bag-Of-Words\\
\hline
3& MOL     & Bag-Of-MOL\\
\hline
4& B+M     & Bag-Of-Words embodying the MOL  \\
\hline
5& BERT \& fastText    & Multilingual pre-trained models \\
\hline
\end{tabular}
\end{table}

\begin{enumerate}
    \item POS+S: we extracted the occurrence of part-of-speech tags for each comment. In addition, we extracted the occurrence of positive and negative words for each comment using the sentiment \cite{sentilexpt2012} and emotion \cite{WordNetAffectBR2008} lexicons. 
    
    \item BOW: we created a bag-of-words representation or, in other words, we generated a text representation that describes the occurrence of dataset vocabulary for each comment. We simply calculate how many times each word of our dataset vocabulary (features) appears in each comment. 
    
    \item MOL: in this representation, a bag-of-words was generated using the terms or expressions extracted from the offensive lexicon (MOL), which were used as features. Therefore, for each comment, the occurrence of the MOL's terms was counted. Additionally, context labels (context-independent and context-dependent) have been considered in order to compute different weights to context-independent and context-dependent features. The frequency of the terms with context-independent labels were multiplied by 2, while the frequency of the terms with context-dependent features remained the same. Specifically for hate speech detection task, we also checked if a term presented any markers that identify hate speech content, and, if this condition was true, an additional weight was accounted. Therefore, in the MOL representation, the value of a term $x$ in the document (comment) $y$ for the offensive comment detection (task 1) is defined according to
    
    \begin{equation}
    \small
        MOL_{x,y} = freq_{x,y} * weightC_{x}
    \end{equation}
    
    and for the hate speech detection (task 2) is given by
    
    \begin{equation}
    \small
        MOL_{x,y} = freq_{x,y} * weightC_{x} * weightH_{x}
    \end{equation}
    
    where $freq$ is the frequency of the term in the document, $weightH = 2$ when the term is a marker that identifies hate speech and $weightH = 1$ otherwise, $weightC = 1$ for context-dependent terms and $weightC = 2$ when the term is context-independent.
     
    \item B+M: we generated a bag-of-words representation, which embodies context label information from the offensive lexicon (MOL). In other words, we firstly generated a bag-of-words from all comments into the dataset. Then, we performed the match with terms into MOL, and then we assigned a weight for terms or expressions labeled with context-dependent (weaker weight) and context-independent (stronger weight). The contextual labels are provided by MOL. Therefore, in B+M representation, the value of a term $x$ into the document (comment) $y$ is defined according to
    
    \begin{equation}
    \small
        B+M_{x,y} = freq_{x,y} * weightC_{x}
    \end{equation}
    
    where $freq$ is the frequency of the term in the document, $weightC = 2$ for context-dependent terms and $weightC = 3$ when the term is context-independent.
    
    \item In a different setting, the feature extraction for the BERT and fastText followed state of the art text classification with a maximum size of 500. For the fastText classifier, we set the maximum size equal to 64 and the maximum number of features equal to 10,000. We used the standard processor model and evaluated the n-gram range for unigram, bigram, and trigram.

\end{enumerate}

%--------------------------------------------------
\subsection{The Learning Methods} 
%--------------------------------------------------
In general, previous works on hate speech detection use neural networks or traditional machine learning techniques on specific communities \cite{davidsonetall2017, AntigoniEtAll2019, VignaEtall2017, DennisEtAll2015, NemanjaetAll2015}. In order to evaluate the performance of neural networks and traditional machine learning techniques, we used the following learning methods: Support Vector Machine (SVM) \cite{scholkopf2001learning} with linear kernel; Multinomial Naive Bayes (NB) \cite{mccallum1998comparison, eyheramendy2003naive}; Multilayer Perceptron (MLP) \cite{haykin2009neural} with one  hidden layer (with 100 neurons), and ReLU activation function; Long Short-Term Memory (LSTM)  \cite{hochreiter1997} with two hidden layers (with 200 and 50 neurons, respectively) and a softmax output unit for the binary classification. ReLU was used as the activation function, as well as number of epochs equal to 10, and a random batch size of 100 documents. Moreover, we also used pre-trained models of word embeddings, such as BERT \cite{bertdevlinetal2019} and fastText \cite{fasttex2016}. %Deep artificial neural networks present a difference from traditional machine learning techniques, mainly due to the learning in multiple layers of abstract features. 

%----------------------------------------------------------
\section{Experiments}
\label{sec:experiments}
%----------------------------------------------------------
We carried out a wide variety of experiments. We describe the entire process in Sections \ref{subsec:data-prep}, \ref{subsec:fs}, \ref{subsec:class-balancing} and \ref{subsec:evaluation}. 

%----------------------------------------------------------
\subsection{Data Preparing} 
\label{subsec:data-prep}
\begin{figure}[!ht]
\centering  
\includegraphics[width=0.50\textwidth]{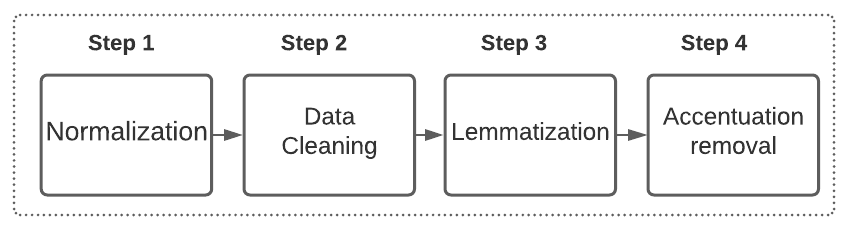}
\caption{Data preparation.}
\label{fig:data-preparing}
\end{figure}

%----------------------------------------------------------
We accomplished an approach for data preparation, as shown in Figure \ref{fig:data-preparing}

Firstly, we normalized our dataset using the normalization tool for Brazilian Portuguese proposed by \citeauthor{costabertagliavolpnunes2016exploring} (\citeyear{costabertagliavolpnunes2016exploring}). The normalization process consists of identifying noise, which is very common in User-Generated Content (UGC), such as orthographic errors, often phonetically-motivated, abbreviations and expressions often used informally by web users, proper names and acronyms wrongly or not at all capitalized, agglutinated words that should be split, and wrong use of sentence delimiters; and suggesting possible substitutions.  

Moving forward, in the second step, we remove emoticons, special characters, accounts, hyperlinks, and websites. In step 3, we lemmatize our dataset using Spacy\footnote{https://spacy.io/models/pt}. Finally, in step 4, accentuation is removed.

%----------------------------------------------------------
\subsection{Feature Selection} 
%----------------------------------------------------------
\label{subsec:fs}
Feature selection (FS) allows the removal of irrelevant and redundant features. In this paper, in order to select the best feature set, we applied the following FS algorithms: (i) Correlation-based Feature Selection (CFS) \cite{hall1998correlation}, which selects characteristics that are highly correlated with the class and not correlated with each other using Pearson coefficient\footnote{Pearson's correlation coefficient is a linear correlation coefficient that returns a value between -1 and +1. A -1 means there is a strong negative relationship, and +1 means there is a strong positive relationship.} as criteria, and (ii) Information Gain Analysis (InfoGain) \cite{witten2016data}, which quantifies and chooses the characteristics that have the maximum information gain concerning the class. We apply both FS techniques on the NB, SVM, MLP and LSTM models. For BERT and fastText features, we do not apply FS techniques. Finally, we evaluated the performance of the FS techniques for each feature representation. More specifically, we measure the potential of the algorithms to help in the gain and loss of accuracy, precision, recall, and f1-score. Results are shown in Table \ref{tab:fstask1}.

%----------------------------------------------------------
\subsection{Class Balancing} 
\label{subsec:class-balancing}
%----------------------------------------------------------
The most common class balancing methods are oversampling \cite{chawla2002smote} and undersampling \cite{witten2016data}. In undersampling, the number of examples of each class is maintained based on the number of examples from the minority class. Differently, in oversampling, the approach involves the construction of examples for the minority class, although these examples may not add any new information to the model. In
our experiments, we adopted the undersampling on the unbalanced classes of hate speech, specially due to the fact that this approach makes overfitting unlikely. Note that, in our dataset (the HateBR), there are 727 labeled hate speech samples versus 2,227 labeled non-hate speech samples. As a result of the undersampling approach, we obtained 727 labeled samples for hate speech and 727 samples for non-hate speech.

%----------------------------------------------------------
\subsection{Evaluation}
\label{subsec:evaluation}
%----------------------------------------------------------
Our models were trained and tested using 10-fold cross-validation \cite{stone1974cross}. We have computed the classical machine learning evaluation measures of Precision, Recall and F1-Score. We present these evaluative measures for each class involved, as well as simple arithmetic means. The results are shown in Table \ref{tab:evaluation}. Moreover, we evaluated BERT and fastText pre-trained models, and show the obtained results in Table \ref{tab:bertfast1}. 

We also present the evaluation of the methods with feature selection (FS). We measure the gain and loss of precision, recall, and f1-score for each selected algorithm (CFS and InfoGain) in both tasks: offensive language detection and hate speech detection, as well as for each representations: POS+S, BOW, MOL and B+M. Table \ref{tab:fstask1} shows the results. We should point out that T1 is the sum of each representation, and T2 is the sum for each FS algorithm.

Finally, Table \ref{tab:c-results} shows the comparison of the results with the current baseline methods for Portuguese.

%------------------------------------------------------
%Evaluation
%------------------------------------------------------
\begin{table*}[!htb]
\scalefont{0.65} %Scale of table font
\caption{NB, SVM, MLP and LSTM Evaluation.}
\centering
\label{tab:evaluation}
\begin{tabular}{@{}l|l|l|c|c|c|c|c|c|c|c|c|c|c|c@{}}
\bottomrule
\multirow{2}{*}{Tasks}                                                                               & \multirow{2}{*}{Features set} & \multirow{2}{*}{Class} & \multicolumn{4}{c|}{Precision}              & \multicolumn{4}{c|}{Recall}                          & \multicolumn{4}{c}{F1-Score}               \\ \cmidrule(l){4-15} 
                                                                                                     &                               &                        & NB            & SVM  & MLP           & LSTM & NB            & SVM           & MLP           & LSTM & NB            & SVM  & MLP           & LSTM \\ \bottomrule
\multirow{12}{*}{\begin{tabular}[c]{@{}l@{}}Task 1:\\ Offensive\\ Language\\ Detection\end{tabular}} & \multirow{3}{*}{POS+S}        & 0                      & 0.50          & 0.51 & 0.47          & 0.49 & 0.41          & 0.39          & 0.51          & 0.37 & 0.45          & 0.44 & 0.49          & 0.42 \\ \cmidrule(l){3-15} 
                                                                                                     &                               & 1                      & 0.50          & 0.51 & 0.54          & 0.49 & 0.50          & 0.64          & 0.51          & 0.62 & 0.59          & 0.57 & 0.52          & 0.55 \\ \cmidrule(l){3-15} 
                                                                                                     &                               & Avg                    & 0.50          & 0.51 & 0.51          & 0.49 & 0.50          & 0.51          & 0.51          & 0.49 & 0.50          & 0.50 & 0.51          & 0.49 \\ \cmidrule(l){2-15} 
                                                                                                     & \multirow{3}{*}{BOW}          & 0                      & 0.85          & 0.82 & 0.92          & 0.83 & 0.86          & 0.96          & 0.81          & 0.89 & 0.86          & 0.88 & 0.81          & 0.86 \\ \cmidrule(l){3-15} 
                                                                                                     &                               & 1                      & 0.86          & 0.95 & 0.79          & 0.88 & 0.85          & 0.79          & 0.90          & 0.81 & 0.85          & 0.86 & 0.90          & 0.85 \\ \cmidrule(l){3-15} 
                                                                                                     &                               & Avg                    & 0.85          & 0.88 & 0.86          & 0.85 & 0.85          & 0.87          & 0.86          & 0.85 & 0.85          & 0.87 & 0.84          & 0.85 \\ \cmidrule(l){2-15} 
                                                                                                     & \multirow{3}{*}{MOL}          & 0                      & 0.74          & 0.78 & 0.94          & 0.79 & 0.97          & 0.96          & 0.77          & 0.94 & 0.84          & 0.86 & 0.85          & 0.86 \\ \cmidrule(l){3-15} 
                                                                                                     &                               & 1                      & 0.95          & 0.94 & 0.72          & 0.93 & 0.66          & 0.73          & 0.93          & 0.75 & 0.78          & 0.82 & 0.81          & 0.83 \\ \cmidrule(l){3-15} 
                                                                                                     &                               & Avg                    & 0.85          & 0.86 & 0.83          & 0.86 & 0.81          & 0.84          & 0.85          & 0.84 & 0.81          & 0.84 & 0.81          & 0.84 \\ \cmidrule(l){2-15} 
                                                                                                     & \multirow{3}{*}{B+M}          & 0                      & 0.84          & 0.84 & 0.91          & 0.86 & 0.93          & 0.94          & 0.83          & 0.85 & 0.88          & 0.88 & 0.87          & 0.85 \\ \cmidrule(l){3-15} 
                                                                                                     &                               & 1                      & 0.93          & 0.93 & 0.81          & 0.85 & 0.83          & 0.81          & 0.90          & 0.86 & 0.88          & 0.87 & 0.86          & 0.85 \\ \cmidrule(l){3-15} 
                                                                                                     &                               & Avg                    & \textbf{0.89} & 0.88 & 0.86          & 0.85 & \textbf{0.88} & \textbf{0.88} & 0.87          & 0.85 & \textbf{0.88} & 0.86 & 0.86          & 0.85 \\ \bottomrule
\multirow{12}{*}{\begin{tabular}[c]{@{}l@{}}Task 2:\\ Hate Speech\\ Detection\end{tabular}}          & \multirow{3}{*}{POS+S}        & 0                      & 0.52          & 0.49 & 0.42          & 0.52 & 0.48          & 0.78          & 0.53          & 0.47 & 0.50          & 0.60 & 0.47          & 0.50 \\ \cmidrule(l){3-15} 
                                                                                                     &                               & 1                      & 0.52          & 0.47 & 0.63          & 0.52 & 0.56          & 0.20          & 0.52          & 0.57 & 0.54          & 0.28 & 0.57          & 0.54 \\ \cmidrule(l){3-15} 
                                                                                                     &                               & Avg                    & 0.52          & 0.48 & 0.53          & 0.52 & 0.52          & 0.49          & 0.53          & 0.52 & 0.52          & 0.44 & 0.52          & 0.52 \\ \cmidrule(l){2-15} 
                                                                                                     & \multirow{3}{*}{BOW}          & 0                      & 0.62          & 0.84 & 0.43          & 0.85 & 0.82          & 0.42          & 0.82          & 0.37 & 0.70          & 0.55 & 0.57          & 0.54 \\ \cmidrule(l){3-15} 
                                                                                                     &                               & 1                      & 0.73          & 0.61 & 0.91          & 0.61 & 0.49          & 0.92          & 0.61          & 0.93 & 0.59          & 0.73 & 0.73          & 0.73 \\ \cmidrule(l){3-15} 
                                                                                                     &                               & Avg                    & 0.68          & 0.72 & 0.67          & 0.73 & 0.66          & 0.67          & 0.72          & 0.66 & 0.65          & 0.64 & 0.65          & 0.64 \\ \cmidrule(l){2-15} 
                                                                                                     & \multirow{3}{*}{MOL}          & 0                      & 0.61          & 0.62 & 0.58          & 0.60 & 0.74          & 0.80          & 0.68          & 0.93 & 0.67          & 0.69 & 0.63          & 0.73 \\ \cmidrule(l){3-15} 
                                                                                                     &                               & 1                      & 0.67          & 0.71 & 0.73          & 0.84 & 0.53          & 0.50          & 0.63          & 0.38 & 0.59          & 0.59 & 0.68          & 0.52 \\ \cmidrule(l){3-15} 
                                                                                                     &                               & Avg                    & 0.64          & 0.66 & 0.66          & 0.72 & 0.64          & 0.65          & 0.66          & 0.65 & 0.63          & 0.64 & 0.66          & 0.63 \\ \cmidrule(l){2-15} 
                                                                                                     & \multirow{3}{*}{B+M}          & 0                      & 0.79          & 0.77 & 0.93          & 0.71 & 0.78          & 0.93          & 0.79          & 0.89 & 0.78          & 0.84 & 0.86          & 0.79 \\ \cmidrule(l){3-15} 
                                                                                                     &                               & 1                      & 0.78          & 0.92 & 0.76          & 0.85 & 0.79          & 0.72          & 0.92          & 0.64 & 0.79          & 0.80 & 0.83          & 0.73 \\ \cmidrule(l){3-15} 
                                                                                                     &                               & Avg                    & 0.78          & 0.84 & \textbf{0.85} & 0.78 & 0.78          & 0.83          & \textbf{0.86} & 0.77 & 0.78          & 0.82 & \textbf{0.85} & 0.76 \\ \bottomrule
\end{tabular}
\end{table*}

%----------------------------------------------------
%BERT and FastText results
%----------------------------------------------------
\begin{table*}[!htb]
\scalefont{0.7374} %Scale of table font
\caption{BERT and fastText Evaluation.}
\label{tab:bertfast1}
\centering
\begin{tabular}{@{}l|l|c|c|c|c|c|c@{}}
\bottomrule
\multirow{2}{*}{Models}             & \multirow{2}{*}{Class} & \multicolumn{3}{c|}{Task 1: Offensive Language Detection} & \multicolumn{3}{c}{Task 2: Hate Speech Detection} \\ \cmidrule(l){3-8} 
                                    &                        & Precision         & Recall            & F1-Score          & Precision       & Recall          & F1-Score       \\ \bottomrule
\multirow{3}{*}{BERT}               & 0                      & 0.85              & 0.86              & 0.86              & 0.76            & 0.65            & 0.70           \\ \cmidrule(l){2-8} 
                                    & 1                      & 0.85              & 0.85              & 0.85              & 0.64            & 0.75            & 0.69           \\ \cmidrule(l){2-8} 
                                    & Avg                    & 0.86              & 0.86              & 0.86              & 0.70            & 0.70            & 0.70           \\ \midrule
\multirow{3}{*}{fastText (unigram)} & 0                      & 0.88              & 0.88              & 0.88              & 0.78            & 0.76            & 0.77           \\ \cmidrule(l){2-8} 
                                    & 1                      & 0.87              & 0.87              & 0.87              & 0.76            & 0.79            & 0.77           \\ \cmidrule(l){2-8} 
                                    & Avg                    & \textbf{0.88}     & \textbf{0.88}     & \textbf{0.88}     & 0.77            & 0.79            & 0.77           \\ \midrule
\multirow{3}{*}{fastText (bigrams)}  & 0                      & 0.83              & 0.87              & 0.85              & 0.77            & 0.84            & 0.80           \\ \cmidrule(l){2-8} 
                                    & 1                      & 0.87              & 0.84              & 0.85              & 0.80            & 0.72            & 0.76           \\ \cmidrule(l){2-8} 
                                    & Avg                    & 0.85              & 0.85              & 0.85              & 0.78            & 0.78            & 0.78           \\ \midrule
\multirow{3}{*}{fastText (trigrams)} & 0                      & 0.83              & 0.91              & 0.87              & 0.77            & 0.97            & 0.86           \\ \cmidrule(l){2-8} 
                                    & 1                      & 0.90              & 0.81              & 0.85              & 0.96            & 0.70            & 0.81           \\ \cmidrule(l){2-8} 
                                    & Avg                    & 0.86              & 0.86              & 0.86              & \textbf{0.86}   & \textbf{0.84}   & \textbf{0.83}  \\ \bottomrule
\end{tabular}
\end{table*}

%----------------------------------------
%Features Selection performance
%-----------------------------------------
\begin{table*}[!htb]
\scalefont{0.62} %Scale of table font
\caption{Feature selection performance.}
\centering
\label{tab:fstask1}
\begin{tabular}{l|l|l|c|c|c|c|c|c|c|c|c|c|c|c}
\hline
                                                                      &                                                                       &                                                                          & \multicolumn{6}{c|}{Task 1: Offensive Language Detection}                                                             & \multicolumn{6}{c}{Task 2: Hate Speech Detection}                                                                    \\ \cline{4-15} 
                                                                      &                                                                       &                                                                          & \multicolumn{4}{c|}{Learning Methods} &                      & {\color[HTML]{333333} }                                & \multicolumn{4}{c|}{Learning Methods} &                      & {\color[HTML]{333333} }                                \\ \cline{4-7} \cline{10-13}
\multirow{-3}{*}{Measures}                                             & \multirow{-3}{*}{FS}                                                  & \multirow{-3}{*}{\begin{tabular}[c]{@{}l@{}}Features\\ set\end{tabular}} & NB      & SVM     & MLP     & LSTM    & \multirow{-2}{*}{T1} & \multirow{-2}{*}{{\color[HTML]{333333} T2}}            & NB      & SVM     & MLP     & LSTM    & \multirow{-2}{*}{T1} & \multirow{-2}{*}{{\color[HTML]{333333} T2}}            \\ \hline
                                                                      & \multicolumn{1}{c|}{}                                                 & POS+S                                                                    & -0.25   & -0.01   & -0.02   & -0.01   & -0.29                & {\color[HTML]{000000} }                                & -0.04   & 0.02    & 0.03    & -0.04   & -0.03                & {\color[HTML]{000000} }                                \\ \cline{3-8} \cline{10-14}
                                                                      & \multicolumn{1}{c|}{}                                                 & BOW                                                                      & -0.01   & -0.02   & 0.00    & -0.02   & -0.05                & {\color[HTML]{000000} }                                & 0.10    & 0.20    & 0.17    & 0.16    & 0.63                 & {\color[HTML]{000000} }                                \\ \cline{3-8} \cline{10-14}
                                                                      & \multicolumn{1}{c|}{}                                                 & MOL                                                                      & -0.04   & -0.02   & -0.04   & -0.04   & -0.14                & {\color[HTML]{000000} }                                & 0.00    & 0.05    & 0.01    & 0.10    & 0.16                 & {\color[HTML]{000000} }                                \\ \cline{3-8} \cline{10-14}
                                                                      & \multicolumn{1}{c|}{\multirow{-4}{*}{CFS}}                            & B+M                                                                      & -0.03   & -0.03   & 0.00    & 0.05    & -0.01                & \multirow{-4}{*}{{\color[HTML]{000000} -0.49}}         & 0.06    & 0.05    & 0.08    & 0.09    & 0.28                 & \multirow{-4}{*}{{\color[HTML]{000000} 1.04}}          \\ \cline{2-15} 
                                                                      &                                                                       & POS+S                                                                    & -0.25   & 0.01    & -0.02   & -0.01   & -0.27                & {\color[HTML]{000000} }                                & -0.02   & 0.01    & 0.04    & -0.01   & 0.02                 & {\color[HTML]{000000} }                                \\ \cline{3-8} \cline{10-14}
                                                                      &                                                                       & BOW                                                                      & 0.00    & 0.02    & 0.05    & 0.07    & 0.14                 & {\color[HTML]{000000} }                                & 0.09    & 0.22    & 0.16    & 0.17    & 0.64                 & {\color[HTML]{000000} }                                \\ \cline{3-8} \cline{10-14}
                                                                      &                                                                       & MOL                                                                      & -0.02   & 0.00    & 0.00    & 0.02    & 0.00                 & {\color[HTML]{000000} }                                & -0.01   & 0.06    & 0.02    & 0.11    & 0.18                 & {\color[HTML]{000000} }                                \\ \cline{3-8} \cline{10-14}
\multirow{-8}{*}{Precision}                                           & \multirow{-4}{*}{\begin{tabular}[c]{@{}l@{}}Info\\ Gain\end{tabular}} & B+M                                                                      & 0.00    & 0.01    & 0.04    & 0.07    & 0.12                 & \multirow{-4}{*}{{\color[HTML]{000000} -0.01}}         & 0.07    & 0.06    & 0.10    & 0.07    & 0.30                 & \multirow{-4}{*}{{\color[HTML]{000000} 1.14}}          \\ \hline
                                                                      & \multicolumn{1}{c|}{}                                                 & POS+S                                                                    & 0.00    & 0.00    & 0.02    & 0.01    & 0.03                 & {\color[HTML]{333333} }                                & -0.04   & 0.01    & 0.02    & -0.04   & -0.05                & {\color[HTML]{000000} }                                \\ \cline{3-8} \cline{10-14}
                                                                      & \multicolumn{1}{c|}{}                                                 & BOW                                                                      & -0.04   & -0.05   & 0.02    & -0.02   & -0.09                & {\color[HTML]{333333} }                                & 0.08    & 0.08    & 0.21    & 0.10    & 0.47                 & {\color[HTML]{000000} }                                \\ \cline{3-8} \cline{10-14}
                                                                      & \multicolumn{1}{c|}{}                                                 & MOL                                                                      & -0.12   & -0.05   & -0.01   & -0.10   & -0.28                & {\color[HTML]{333333} }                                & 0.00    & 0.01    & 0.07    & 0.02    & 0.10                 & {\color[HTML]{000000} }                                \\ \cline{3-8} \cline{10-14}
                                                                      & \multicolumn{1}{c|}{\multirow{-4}{*}{CFS}}                            & B+M                                                                      & -0.05   & -0.05   & 0.02    & 0.05    & -0.03                & \multirow{-4}{*}{{\color[HTML]{333333} -0.37}}         & 0.06    & 0.04    & 0.10    & 0.08    & 0.28                 & \multirow{-4}{*}{{\color[HTML]{000000} 0.80}}          \\ \cline{2-15} 
                                                                      &                                                                       & POS+S                                                                    & 0.00    & 0.00    & -0.02   & -0.01   & -0.03                & {\color[HTML]{000000} }                                & -0.02   & 0.01    & 0.04    & -0.01   & 0.02                 & {\color[HTML]{000000} }                                \\ \cline{3-8} \cline{10-14}
                                                                      &                                                                       & BOW                                                                      & 0.00    & 0.02    & 0.05    & 0.07    & 0.14                 & {\color[HTML]{000000} }                                & -0.02   & -0,01   & 0.04    & -0.02   & -0.01                & {\color[HTML]{000000} }                                \\ \cline{3-8} \cline{10-14}
                                                                      &                                                                       & MOL                                                                      & -0.04   & -0.01   & 0.01    & -0.04   & -0.08                & {\color[HTML]{000000} }                                & -0.02   & 0.02    & 0.08    & 0.04    & 0.12                 & {\color[HTML]{000000} }                                \\ \cline{3-8} \cline{10-14}
\multirow{-8}{*}{Recall}                                              & \multirow{-4}{*}{\begin{tabular}[c]{@{}l@{}}Info\\ Gain\end{tabular}} & B+M                                                                      & 0.00    & 0.01    & 0.04    & 0.08    & 0.13                 & \multirow{-4}{*}{{\color[HTML]{000000} 0.16}}          & 0.07    & 0.05    & 0.11    & 0.00    & 0.23                 & \multirow{-4}{*}{{\color[HTML]{000000} 0.36}}          \\ \hline
                                                                      & \multicolumn{1}{c|}{}                                                 & POS+S                                                                    & -0.17   & -0.03   & 0.04    & 0.02    & -0.14                & {\color[HTML]{000000} }                                & -0.04   & -0.06   & 0.03    & -0.04   & -0.11                & {\color[HTML]{333333} }                                \\ \cline{3-8} \cline{10-14}
                                                                      & \multicolumn{1}{c|}{}                                                 & BOW                                                                      & -0.04   & -0.05   & -0.07   & -0.02   & -0.13                & {\color[HTML]{000000} }                                & 0.07    & 0.12    & 0.15    & 0.07    & 0.41                 & {\color[HTML]{333333} }                                \\ \cline{3-8} \cline{10-14}
                                                                      & \multicolumn{1}{c|}{}                                                 & MOL                                                                      & -0.14   & -0.06   & -0.02   & -0.12   & -0.39                & {\color[HTML]{000000} }                                & -0.01   & -0.01   & -0.02   & -0.03   & -0.07                & {\color[HTML]{333333} }                                \\ \cline{3-8} \cline{10-14}
                                                                      & \multicolumn{1}{c|}{\multirow{-4}{*}{CFS}}                            & B+M                                                                      & -0,06   & -0.05   & -0.02   & 0.05    & -0.08                & \multirow{-4}{*}{{\color[HTML]{000000} -0.74}}         & 0.06    & 0.04    & 0.08    & 0.07    & 0.25                 & \multirow{-4}{*}{{\color[HTML]{333333} \textbf{0.48}}} \\ \cline{2-15} 
                                                                      &                                                                       & POS+S                                                                    & -0.17   & 0.11    & -0.01   & 0.03    & -0.04                & {\color[HTML]{333333} }                                & -0.07   & -0.02   & 0.03    & -0.01   & -0.07                & {\color[HTML]{333333} }                                \\ \cline{3-8} \cline{10-14}
                                                                      &                                                                       & BOW                                                                      & 0.00    & 0.02    & -0.01   & 0.07    & 0.14                 & {\color[HTML]{333333} }                                & 0.07    & -0.02   & -0.01   & -0.04   & -0.09                & {\color[HTML]{333333} }                                \\ \cline{3-8} \cline{10-14}
                                                                      &                                                                       & MOL                                                                      & -0.05   & -0.01   & 0.05    & -0.04   & -0.11                & {\color[HTML]{333333} }                                & -0.01   & 0.00    & -0.01   & 0.02    & -0.01                & {\color[HTML]{333333} }                                \\ \cline{3-8} \cline{10-14}
\multirow{-8}{*}{\begin{tabular}[c]{@{}l@{}}F1-\\ Score\end{tabular}} & \multirow{-4}{*}{\begin{tabular}[c]{@{}l@{}}Info\\ Gain\end{tabular}} & B+M                                                                      & 0.00    & 0.01    & 0.04    & 0.08    & 0.13                 & \multirow{-4}{*}{{\color[HTML]{333333} \textbf{0.12}}} & 0.06    & 0.05    & 0.10    & -0.02   & 0.20                 & \multirow{-4}{*}{{\color[HTML]{333333} 0.03}}          \\ \hline
\end{tabular}
\end{table*}

%----------------------Results-----------------------
\section{Results}
\label{subsec:results}
%----------------------------------------------------
As shown in Table \ref{tab:evaluation}, the B+M proposed method in this paper obtained better results of precision, recall, and f1-score in both tasks - offensive language and hate speech detection. The worse results were obtained using the POS+S approach, which combines part-of-speech and sentiment lexicon features. We should point out the considerable impact of an offensive lexicon for abusive language detection, when compared to the impact of a sentiment lexicon. Our results showed that sentiment lexicon approach present weak performance for abusive language detection on the web and social media.

Moving forward, the conducted experiments also show that the traditional machine learning techniques presented better performance than neural-based classifiers for offensive language tasks. Nevertheless, for the hate speech detection task, the neural-based classifier overcame the traditional machine learning methods. 

In general, BERT and fastText, as shown in Table \ref{tab:bertfast1}, presented a high performance for both tasks (offensive language and hate speech detection), even though our approach (B+M) has overcome the fastText (trigrams) in 2\% (f1-score) for hate speech detection, as well as presented better precision performance, and the same recall and f1-score performances for offensive language detection. 

Considering the feature selection (FS) performance, as shown in Table \ref{tab:fstask1}, the InfoGain algorithm produced better results for precision, recall, and f1-score than CFS algorithm for offensive language detection (task 1). On other hand, for hate speech detection (task 2),  CFS algorithm obtained better performance than InfoGain in recall and f1-score.  Moreover, for offensive language detection (task 1), InfoGain applied on BOW and B+M representations obtained performance gain, and POS+S and MOL presented loss of performance. For hate speech detection (task 2), InfoGain applied on B+M representation presented performance gain. Differently from this, POS+S, BOW, and MOL had a loss of performance using InfoGain. Differently, CFS algorithm applied to BOW and B+M obtained performance gain, and when applied to POS+S and MOL representations, presented loss of performance.

%-----------------------------------------------------
\subsection{Comparing Results}
%----------------------------------------------------
Table \ref{tab:c-results} shows a comparison of results between our new proposed method and baseline methods for Portuguese. Although a direct comparison is unfair (as the authors use different datasets), it offers an idea of the general performance of the methods.

\citeauthor{MoreiraAndPelle2017} (\citeyear{MoreiraAndPelle2017}) report a f1-score of 81\% using SVM and NB algorithms. For the same algorithms, our approach presented 88\% of f1-score, improving the performance. In the same settings, \citeauthor{fortuna2019hierarchically} (\citeyear{fortuna2019hierarchically}) report a f1-score of 78\% using the LSTM algorithm. In our experiments, we obtained an f1-score of 86\%, also using the LSTM algorithm, consequently, our approach presented better performance.

\begin{table}[!htb]
\scalefont{0.60}
\caption{Comparison of results.}
\label{tab:c-results}
\begin{tabular}{llll} \hline 
 & \multicolumn{1}{c}{\begin{tabular}[c]{@{}c@{}}\textbf{Dataset} \\ \textbf{language}\end{tabular}} & \multicolumn{1}{c}{\textbf{Algorithms}} & \multicolumn{1}{c}{\textbf{F1-score}} \\ \hline  
 
Our approach & \begin{tabular}[c]{@{}l@{}}Brazilian \\ Portuguese\end{tabular} & SVM and NB & \textbf{88\%} \\ \hline
\citeauthor{MoreiraAndPelle2017} (\citeyear{MoreiraAndPelle2017}) & \begin{tabular}[c]{@{}l@{}}Brazilian \\ Portuguese\end{tabular} & SVM and NB & 81\% \\ \hline \hline

Our approach & \begin{tabular}[c]{@{}l@{}}Brazilian \\ Portuguese\end{tabular} & LSTM & \textbf{86\%} \\ \hline
\citeauthor{fortuna2019hierarchically} (\citeyear{fortuna2019hierarchically}) & \begin{tabular}[c]{@{}l@{}}European and \\Brazilian\\ Portuguese\end{tabular} & LSTM & 78\% \\ \hline
\end{tabular}
\end{table}

%--------------------------------------------------------------------
\section{Conclusions} 
\label{sec:conclusions}
%--------------------------------------------------------------------
In this work, we provide a new approach for the automatic detection of abusive comments on social media. Our approach embodies an offensive lexicon that provides contextual information. Due to the increase of abusive comments on social media in Brazil, as well as the lack of research in Portuguese, we decided to use an Brazilian annotated dataset to evaluate the models. The proposed approach obtains high performances: 88\% f1-score for offensive comments detection, and 85\% for comments with hate speech, which overcame the current baseline methods for Portuguese. We also evaluated the performance of feature selection (FS) methods, and conclude that InfoGain algorithm is the best algorithm for the offensive comment detection task, considering the obtained gains in recall and f1-score. For the hate speech detection task, CFS algorithm obtained better performance. Accordingly, based on the obtained results, we concluded that the proposed approach in this paper for automated detection of abusive comments is efficient and highly relevant, bearing in mind the current Brazilian social scenario, in which hateful comments are a very relevant social problem. Moreover, in the next year (2022), there will be presidential elections in Brazil, and this paper may provide a reliable automated approach for abusive comments detection in order to minimize political polarization, as well as hate crimes on social media.

%The practice is known as cyberbullying \footnote{Cyberbullying is a type of systematic violence in cyberspace, which to be represented by several authors as a public health issue \cite{Brownetall2006}}, and defamation \footnote{According to article 139 of the Brazilian Penal Code, defamation consists of imputing to someone an offense to their reputation.}, which are behaviors that may lead to lawsuits criminal consequences. 

%Moreover, in the next year (2022), there will be presidential elections in Brazil, and this paper may provide a reliable automated approach for abusive comments detection in order to minimize political polarization and hate crimes on social media, as well as to protect the Brazilian population.

%---------------------------------------------
\section*{Acknowledgements}
%---------------------------------------------
The authors are grateful to DCC-UFMG research project: WHATSAPP MONITOR - SIMP 18048 for supporting this work.
%---------------------------------------------
%References
%---------------------------------------------
%\citep{goodman-etal-2016-noise} 
\bibliographystyle{acl_natbib}
\bibliography{anthology,ranlp2021}

%\appendix

\end{document}